\pdfoutput=1

\documentclass[11pt]{article}

\usepackage[preprint]{acl}

\usepackage{times}
\usepackage{latexsym}

\usepackage[T1]{fontenc}

\usepackage[utf8]{inputenc}

\usepackage{amsmath}
\usepackage{amssymb}
\usepackage{amsthm}
\newtheorem{definition}{Definition}

\newtheorem{theorem}{Theorem}

\usepackage{multirow}
\usepackage{makecell}
\usepackage{algorithm}
\usepackage{algorithmic}

\usepackage{microtype}

\usepackage{inconsolata}

\usepackage{graphicx}
\usepackage{subfloat}
\usepackage{subcaption}
\title{SC-LoRA: Balancing Efficient Fine-tuning and Knowledge Preservation via Subspace-Constrained LoRA}

\renewcommand\footnotemark{}
\author{
\!Minrui Luo$^{*1, 2}$, Fuhang Kuang$^{*1}$, Yu Wang$^{3}$, Zirui Liu$^{1}$, Tianxing He$^{\dagger 2,1,4}$\thanks{$^*$Equal contribution.}\thanks{$^\dagger$Tianxing He is the corresponding author.}\\
$^{1}$Shanghai Qi Zhi Institute\\
  $^{2}$Institute for Interdisciplinary Information Sciences, Tsinghua University\\
  $^{3}$Institute of Information Engineering, Chinese Academy of Sciences\\
  $^{4}$Xiongan AI Institute\\
    \footnotesize{\texttt{\{luomr22,kfh22,liu-zr22\}@mails.tsinghua.edu.cn;}}\\
    \footnotesize{\texttt{wangyu2002@iie.ac.cn}}\quad \footnotesize{\texttt{hetianxing@mail.tsinghua.edu.cn}}\\
}

\begin{document}

\maketitle

\begin{abstract}
Parameter-Efficient Fine-Tuning (PEFT) methods, particularly Low-Rank Adaptation (LoRA), are indispensable for efficiently customizing Large Language Models (LLMs). However, vanilla LoRA suffers from slow convergence speed and knowledge forgetting problems. Recent studies have leveraged the power of designed LoRA initialization, to enhance the fine-tuning efficiency, or to preserve knowledge in the pre-trained LLM. However, none of these works can address the two cases at the same time. To this end, we introduce \textbf{S}ubspace-\textbf{C}onstrained LoRA (\textbf{SC-LoRA}), a novel LoRA initialization framework engineered to navigate the trade-off between efficient fine-tuning and knowledge preservation. We achieve this by constraining the output of trainable LoRA adapters in a low-rank subspace, where the context information of fine-tuning data is most preserved while the context information of preserved knowledge is least retained, in a balanced way. Such constraint enables the trainable weights to primarily focus on the main features of fine-tuning data while avoiding damaging the preserved knowledge features. We provide theoretical analysis on our method, and  conduct extensive experiments including \textit{safety preservation} and \textit{world knowledge preservation}, on various downstream tasks. In our experiments, SC-LoRA succeeds in delivering superior fine-tuning performance while markedly diminishing knowledge forgetting, surpassing contemporary LoRA initialization methods.
\end{abstract}

\section{Introduction}
\label{sec:introduction}

Fine-tuning effectively adapts large language models to downstream tasks \citep{luo2025wizardmath, yu2024metamath}. Due to the high computational cost of full fine-tuning, parameter-efficient fine-tuning (PEFT) methods \citep{xu2023parameterefficientfinetuningmethodspretrained, han2024parameterefficient} have been proposed to reduce the number of trainable parameters while maintaining good fine-tuning performance. Among various PEFT methods, LoRA \citep{hu2022lora} is a simple yet efficient approach that introduces trainable low-rank adaptation modules for tuning. 
While LoRA offers significant parameter efficiency, it has two important problems: (1) the convergence speed of the fine-tuning process is relatively slow due to the noise and zero initialization of adapter modules; (2) it potentially leads to catastrophic forgetting problem \citep{goodfellow2015empiricalinvestigationcatastrophicforgetting} as other fine-tuning methods do, such as harming the world knowledge stored in pre-trained LLMs \citep{CorDA}, and degrading the safety of aligned LLMs \citep{benign_finetune_attack_hex_phi}. 

Recent works have found that carefully designed initialization on LoRA adapters can solve these problems. \citet{PiSSA} initializes LoRA adapters by parts of Singular Value Decomposition (SVD) of original weight $W_0$, leading to faster convergence and improved performance by encapsulating the most significant information stored in $W_0$. Later works \citep{CorDA, EVA} initialize LoRA weights based on semantic information stored in the activations of each layer on the target fine-tuning dataset. These data-driven approaches successfully enhance the fine-tuning speed and performance. Towards catastrophic forgetting problem in LoRA fine-tuning, \citet{CorDA} proposes to initialize LoRA weights by the least principal directions of world knowledge data features, successfully alleviating the forgetting problem. However, these works can only solve either side of the two problems, but do not consider the trade-off between enhancing fine-tuning performance and preserving pre-trained knowledge, which is a common need when doing parameter-efficient fine-tuning.

In this paper, we introduce \textbf{S}ubspace-\textbf{C}onstrained \textbf{LoRA}, a balanced LoRA scheme that achieves both better fine-tuning results and good preservation of knowledge in LLMs. Specifically, we compute directions of linear layer output that align with the principal directions of fine-tuning data and at the same time are orthogonal to the principal directions of preserved knowledge. These directions are then used to initialize the adapter weights, constraining the output vectors (of each adapter layer) in a subspace spanned by these directions. This constraint intuitively makes the updating terms to focus on the fine-tuning data information, while avoiding affecting the preserved knowledge. 
By extensive experiments, we verify that by such constraint on balanced directions, our method achieves both efficient fine-tuning and excellent knowledge preservation, solving the problems that previous methods cannot address. In conclusion, our contribution includes:

\begin{enumerate}
    \item We propose SC-LoRA, a balanced LoRA scheme that can achieve efficient fine-tuning and knowledge preservation at the same time, which previous methods cannot handle.
    \item We provide theoretical proofs to explain our strategies, including analysis on subspace selection and initialization setting.
    \item We conduct extensive experiments regarding both \textit{safety preservation} and \textit{world knowledge preservation} on various downstream tasks, verifying the effectiveness of our method.
\end{enumerate}

\section{Related Work}
\label{sec:related work}

\paragraph{Parameter-Efficient Fine-Tuning (PEFT). } Modern large language models (LLMs) with billions of parameters face significant computational and memory challenges during full-parameter fine-tuning on downstream tasks, motivating the development of Parameter-Efficient Fine-Tuning (PEFT) methods that optimize only a small amount of parameters while maintaining model performance \citep{xu2023parameterefficientfinetuningmethodspretrained, han2024parameterefficient}. 

Common PEFT approaches include partial fine-tuning \citep{ben-zaken-etal-2022-bitfit, pmlr-v235-bu24c} that only tune part of the parameters; soft prompt finetuning \citep{hambardzumyan-etal-2021-warp,  lester-etal-2021-power}, where trainable prompts are appended to inputs with model parameters frozen; adapter tuning \citep{pmlr-v97-houlsby19a,lin-etal-2020-exploring,ruckle-etal-2021-adapterdrop,NEURIPS2021_081be9fd,pfeiffer-etal-2021-adapterfusion,he2022towards,wang-etal-2022-adamix,NEURIPS2023_19d7204a} which inserts additional trainable layers into LLMs and fix the base model parameters; and LoRA\citep{hu2022lora, aghajanyan-etal-2021-intrinsic}, which decomposes weight updates into low-rank matrices. Different from other approaches, LoRA does not change the original model architecture or incurring extra computational cost during inference since the extra adapters can be merged into original parameters.

\paragraph{LoRA Initialization.} Multiple LoRA initialization methods have been proposed, with the aim of improving training efficiency or obtaining other abilities. 

PiSSA \citep{PiSSA} argued that the default initialization of ``Gaussian noise \citep{He_2015_ICCV} and zero'' to the adapters can lead to slow convergence. Hence they propose to apply singular value decomposition to original weight matrices and utilizes the top components to initialize LoRA, encapsulating the most significant information stored in original weights. CorDA \citep{CorDA} utilizes covariance matrices of data context, and takes the first (or last) singular vectors after context-oriented decomposition as initialization of LoRA adapters. They propose two different modes, one for improving fine-tuning performance and the other for mitigating world knowledge forgetting. Similar to CorDA, EVA \citep{EVA} feeds fine-tuning data into the model, applies sigular value decomposition to activation covariance matrices, and takes top singular vectors as initialization weights. LoRA-GA \citep{LoRA-GA} also utilizes data context but applies decomposition on the gradient. \citet{LoRA-Init-BA-which-Zero} analyze the initialization of LoRA adapters, and have shown how the asymmetry of two low rank matrices affects training dynamics.

\paragraph{Harmful Finetuning Attack and Defense strategies.} To prevent potential misuse, LLMs usually undergo specific training to align them with human values before deployment \citep{ouyang2022training, bai2022training}. Nevertheless, jailbreak attacks employ carefully designed inputs to circumvent this alignment, with prominent methods including Greedy Coordinate Gradient (GCG) \citep{zou2023universal}, AutoDAN \citep{liu2023autodan}, and PAIR \citep{chao2023jailbreaking}. Beyond these direct attacks, fine-tuning can also undermine a model's safety alignment, even when non-harmful data is used \citep{benign_finetune_attack_hex_phi, Select_Benign_Data_breaks_safety}. 

Consequently, researchers have developed various defense strategies against such fine-tuning risks, generally falling into following approaches: enhancing the original safety alignment \citep{huang2024vaccine, huang2024booster, SaLoRA}, restricting the gradient of fine-tuning parameters or the scope of trained residuals \citep{pmlr-v235-wei24f, li2025safety}, mixing additional safety data \citep{wang2024backdooralign, huang2024lazy}, modifying the loss function \citep{shallow_deep} and post-fine-tuning processing \citep{xin2024realingment, hsu2024safelora}. 
Different from previous works, our method focuses on an alternative approach of mitigating safety risks during fine-tuning by only modifying initialization, without mixing safety data during fine-tuning, appending prefix during inference time, or adding extra high-rank modules to the model - which would incur computation overhead either in training or inference time.

\paragraph{World Knowledge Forgetting.}

Catastrophic forgetting \citep{mccloskey1989catastrophic} is a phenomenon when models lose previously acquired knowledge when adapting to new tasks, and has been extensively studied in deep learning. Early approaches to solve the problem include knowledge distillation \citep{8107520}, rehearsal \citep{riemer2018learning} and dynamic architectures \citep{Yan_2021_CVPR}. For large language models, preserving world knowledge remains challenging due to massive pre-training data and model size. Recent efforts mitigate forgetting by freezing pre-trained layers while introducing new adapters \citep{wu-etal-2024-llama, dou-etal-2024-loramoe}. Recently \citet{CorDA} proposed CorDA with Knowledge-Preserved Adaptation (KPA) mode, addressing world knowledge forgetting through LoRA initialization. 

\section{Method}
\label{sec:method} 

Below, we first review the vanilla LoRA, and describe our proposed SC-LoRA method.

\begin{figure*}[t]
\centering
\begin{subfigure}{0.49\linewidth}
    \includegraphics[width=\linewidth]{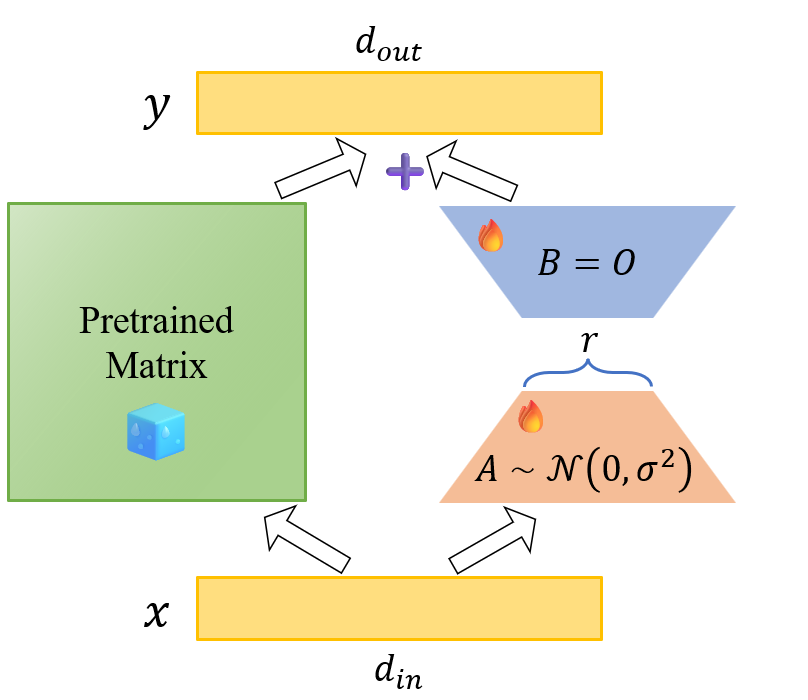} 
    \caption{LoRA}
    \label{fig:teaser-a}
\end{subfigure}
\begin{subfigure}{0.49\linewidth}
    \includegraphics[width=\linewidth]{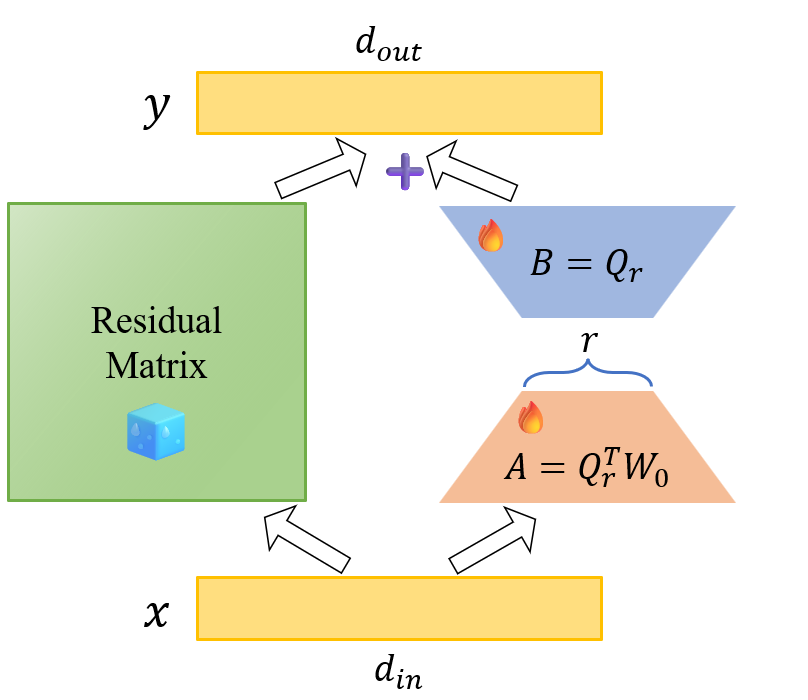}
    \caption{SC-LoRA}
    \label{fig:teaser-b}
\end{subfigure}

\caption{Comparison of LoRA with default Kaiming initialization and our proposed SC-LoRA. (a) LoRA initializes down-projection matrix $A$ by Gaussian noise and up-projection matrix $B$ by zero matrix. (b) Our SC-LoRA initializes $A$ by $Q_r^\top W_0$ and B by $Q_r$, where $Q_r$ consists of $r$ orthonormal vectors as columns obtained by Algorithm \ref{alg:init-pseudocode}.}

\label{fig:teaser}
\end{figure*}

\subsection{LoRA}

 Following the hypothesis that the update of weight matrices presents a low rank structure \citep{aghajanyan-etal-2021-intrinsic}, LoRA \citep{hu2022lora} uses the product of two trainable low-rank matrices to learn the weight change while keeping the original weight matrices frozen. To express in mathematical form, LoRA adds low-rank adapters $A,B$ to original weight matrix $W_0$ by $W^\prime = W_0 + BA$, where $W^\prime, W_0\in \mathbb{R}^{d_{\mathrm{out}} \times d_{\mathrm{in}}}$, $A \in \mathbb{R}^{r \times d_{\mathrm{in}}}$, $B \in \mathbb{R}^{d_{\mathrm{out}} \times r}$, $r\ll \min(d_{\mathrm{in}},d_{\mathrm{out}})$. When fine-tuning, $W_0$ is kept frozen, and $A,B$ are trainable parameters. 

From the default initialization scheme of LoRA, $A$ is initialized by Kaiming Initialization \citep{He_2015_ICCV} while $B$ is initialized by zero matrix. Consequently, the adapter term $BA = O$ and $W^\prime = W_0$ at the start of fine-tuning, ensuring the coherence with the model before fine-tuning. For initializations with non-zero adapter $BA$ \citep{PiSSA, CorDA, LoRA-GA}, the frozen weights are adjusted to the residual term $W_{\mathrm{res}} = W_0 - B_{\mathrm{init}} A_{\mathrm{init}}$. Then the adapted weight is $W^\prime = W_0 - B_{\mathrm{init}} A_{\mathrm{init}} + BA = W_{\mathrm{res}} + BA$. In transformer-based LLMs, LoRA adapters are applied to weight matrices within the self-attention and multilayer perceptron (MLP) layers.

\subsection{SC-LoRA}
\label{sec:SC-LoRA}

Known as catastrophic forgetting problem \cite{chen-etal-2020-recall}, a large language model often performs worse on its pre-trained knowledge after fine-tuning on a downstream task. To this end, we consider fine-tuning a large language model on downstream task $T_+$, while preserving its ability on the other task $T_-$. Consider the output of a linear layer $h=W_0x = W_{\mathrm{res}}x + B_{\mathrm{init}}A_{\mathrm{init}}x$. We denote $\mathcal{P}_+$ and $\mathcal{P}_-$ the distribution of $h$ when the model is fed with data from $T_+$ and $T_-$, respectively. Our aim is to initialize $A, B$ within the $r$-rank constraint so that $BAx$ preserves the most of $\mathcal{P}_+$ and the least of $\mathcal{P}_-$, so that after initialization, the trainable term $BAx$ is constrained to primarily focus on $\mathcal{P}_+$ while avoiding modifying $\mathcal{P}_-$. This is equivalent to identify a low-dimensional subspace $S\subset \mathbb{R}^{d_{\mathrm{out}} }$ with rank $r$, on which the projection of $\mathcal{P}_+$ is mostly preserved and the projection of $\mathcal{P}_-$ is mostly eliminated. To evaluate such property of subspace $S$, we define the following projection $\Pi_S$ and reward $R(S)$:
\begin{definition}
    Suppose $S$ is a subspace of $\mathbb{R}^n$ of dimension $r$, and let $\{q_i\}_{i\in[r]}$ be an orthonormal basis of $S$,
    then the orthogonal projection operator onto $S$, denoted $\Pi_S$, is defined as:
    \begin{equation}
      \begin{aligned}
        \Pi_S(x) &= \sum_{i=1}^r (q_i^\top x) q_i \\
        &= \sum_{i=1}^r (q_iq_i^\top) x, \forall x \in \mathbb{R}^n.
      \end{aligned}
    \end{equation}
    Note: the selection of the orthonormal basis does not affect $\Pi_S$. 
\end{definition}
\begin{definition}
For a subspace $S \subset \mathbb{R}^{d_{\mathrm{out}}}$ of dimension $r$, define the reward $R(S)$ over $\mathcal{P}_\pm$ as: 
\begin{equation}
\label{eq:reward}
  \begin{aligned}
    R(S) &= (1-\beta)\mathbb{E}_{X_{+} \sim \mathcal{P}_{+}} \left[ \left\| \Pi_S\left(X_+\right) \right\|_2^2 \right] \\ &- \beta \mathbb{E}_{X_{-} \sim \mathcal{P}_{-}} \left[ \left\|\Pi_S(X_-)\right\|_2^2 \right],
  \end{aligned}
\end{equation} 
where $\beta \in [0,1]$ is a hyperparameter to tune. 
\end{definition}
The first term of $R(S)$ quantifies the context information of $T_+$ contained in subspace $S$, while the second penalizes that of $T_-$. We use $\beta$ to balance the trade-off between focusing on $T_+$ and preservation on $T_-$. Given the objective to maximize $R(S)$, in the following we provide Theorem \ref{thm:reward} to compute the optimal subspace and then use it to set our LoRA initialization scheme. 

\begin{theorem}\label{thm:reward}
Let $\mathrm{Cov}_+, \mathrm{Cov}_-$ be the covariance matrices of random vectors $X_+\sim\mathcal{P}_+$ and $X_-\sim\mathcal{P}_-$, respectively:
\begin{align}\label{eq:cov}
    \mathrm{Cov}_+ &= \mathbb{E}\left[ X_+ X_+^\top \right], \\
    \mathrm{Cov}_- &= \mathbb{E}\left[ X_- X_-^\top \right].
\end{align}
And let
\begin{equation}
    \Delta \mathrm{Cov} = (1-\beta) \mathrm{Cov}_+ - \beta \mathrm{Cov}_-.
    \label{eq:delta-cov}
\end{equation}
Then do eigenvalue decomposition of $\Delta\mathrm{Cov}$ and take the first $r$ eigenvectors $\{q_i\}_{i\in [r]}$ with the largest eigenvalues.
Then, if following condition holds, the reward $R(S)$ is maximized: 
\begin{equation}
  S = \mathrm{span}\left(\left\{ q_i \right\}_{i\in[r]}\right).
\end{equation}

\begin{proof}
    See Appendix \ref{proof:reward}. 
\end{proof}
\end{theorem}

Theorem \ref{thm:reward} shows the steps to compute the optimal subspace that maximized $R(S)$. Then, to constrain the updating output term $BAx$ in the subspace $S$, we propose our LoRA initialization method:
\begin{align}
    B_{\mathrm{init}} &= (q_1\,q_2\,\cdots\,q_r), \\ \label{eq:sclora-init-B}
    A_{\mathrm{init}} &= (q_1\,q_2\,\cdots\,q_r)^\top W_0, \\ \label{eq:sclora-init-A}
    W_{\mathrm{res}} &= W_0 - B_{\mathrm{init}} A_{\mathrm{init}},
\end{align}
as illustrated in Figure \ref{fig:teaser-b}. To explain the initialization setting, we provide the following theorem:

\begin{theorem}
    \label{thm:init-of-BA}
    Let $h, x$ be the output and input of the original linear layer $W_0$, satisfying $h=W_0x$. When $A, B$ are initialized by Equations \ref{eq:sclora-init-B}, \ref{eq:sclora-init-A}, the following property holds:
    \begin{equation}
        B_{\mathrm{init}}A_{\mathrm{init}}x = \Pi_S(h) \in S,\,\forall  x\in\mathbb{R}^{d_{\mathrm{in}}}.
    \end{equation}
    \begin{proof}
        See Appendix \ref{proof:init-of-BA}.
    \end{proof}
\end{theorem}

Together with Theorem \ref{thm:reward}, our initialization method has the following properties: When $\beta=0$ and the model is fed with data from task $T_+$, $h$ follows distribution $\mathcal{P}_+$, then the norm of the updating term $BAx$ is maximized, providing the most context information of $T_+$ for training; When $\beta=1$ and the model is fed with data from task $T_-$, $h$ follows distribution $\mathcal{P}_-$, then the norm of $BAx$ is minimized, passing the least context information of $T_-$ to trainable parameters. When $\beta\in(0, 1)$, it is the balance between the two cases. The property indicates that, during fine-tuning, the trainable weights are updating more on features related to $T_+$ and less on features related to $T_-$, and hence enhancing learning $T_+$ while avoiding damaging information related to $T_-$.

The pseudo-code of our initialization algorithm is shown in Algorithm \ref{alg:init-pseudocode}. In practice, it is hard to format the true distribution and covariance of output vectors, so we approximate them by feeding hundreds of samples into the model, and use the collection of output vectors to approximate the distribution.

\begin{algorithm}
\caption{SC-LoRA initialization.}
\label{alg:init-pseudocode}
    \begin{algorithmic}[1]
        \REQUIRE Datasets $\mathcal{D}_+, \mathcal{D}_-$ from tasks $T_+, T_-$, respectively.
        \STATE Let $B_+=|\mathcal{D}_+|, B_-=|\mathcal{D}_-|$, $L$ be the length of each sample (clipped to same length).
        \STATE Separately feed samples in $\mathcal{D}_+, \mathcal{D}_-$ into the pre-trained model, collect batched output $\hat{X}_+\in\mathbb{R}^{d_{out}\times B_+L}, \hat{X}_-\in\mathbb{R}^{d_{out}\times B_-L}$ of each linear layer. Within each sample, the output vector is summed over all tokens.
        \STATE $\mathrm{Cov}_+ \leftarrow \frac{1}{B_+} \hat{X}_+ \hat{X}_+^\top$.
        \STATE $\mathrm{Cov}_- \leftarrow \frac{1}{B_-} \hat{X}_- \hat{X}_-^\top$.
        \STATE Do eigenvalue decomposition on $\Delta \mathrm{Cov} = (1-\beta) \mathrm{Cov}_+ - \beta \mathrm{Cov}_-$, and take the first $r$ eigenvectors $\{q_i\}_{i\in r}$ with the largest eigenvalues.
        \STATE $Q_r\leftarrow (q_1\,q_2\,\cdots\,q_r)$.
        \STATE $B_{\mathrm{init}} \leftarrow Q_r$.
        \STATE $A_{\mathrm{init}} \leftarrow Q_r^\top W_0$ .
        \STATE $W_{\mathrm{res}} \leftarrow W_0 - B_{\mathrm{init}} A_{\mathrm{init}}$.
    \end{algorithmic}
\end{algorithm}

\section{Experiments}
\label{sec:experiments}

In the experiments below, we compare SC-LoRA with 5 baselines: 

(1) Full fine-tuning. Fine-tune on all parameters of the model;

(2) Vanilla LoRA \citep{hu2022lora}. Fine-tune only on LoRA adapters, with $B$ initialized with Gaussian noise \citep{He_2015_ICCV}, and $A$ initialized by zero; 

(3) PiSSA \citep{PiSSA}, for efficient fine-tuning. It applies SVD on pre-trained weight $W_0$ and initializes LoRA adapters by the main parts of decomposition; 

(4) CorDA \citep{CorDA} Instruction-Previewed Adaptation (IPA) mode, for efficient fine-tuning. It feeds fine-tuning data into the model to get the covariance of activations, applies self-defined context-oriented decomposition, and initializes LoRA adapters with principal directions obtained in decomposition; 

(5) CorDA Knowledge-Preserved Adaptation (KPA) mode, for knowledge preservation. The initialization algorithm is basically the same as IPA mode except that it feeds preserved knowledge data and take the least principal directions for initialization. 

For initialization of CorDA IPA and KPA mode, we calculate the covariance matrices with 256 samples from fine-tuning dataset and preserved knowledge dataset, respectively with 256 samples. We use AdamW optimizer \citep{loshchilov2019decoupledweightdecayregularization} with the following hyper-parameters: batch size 128, learning rate 2e-5 (except for experiment in Section \ref{subsec: Safety Preservation on Data Poisoning Attack}, where we tune the learning rate of baselines for better performance), cosine annealing learning rate schedule, warm-up ratio 0.03, and no weight decay. The rank of LoRA and its variants are all set to 128 for comparison. For SC-LoRA, we tune the hyperparameter $\beta$ to find a good balanced result. All experiment results are obtained by running on only one seed.

Below we discuss results in three settings: (1) Preservation of world knowledge when fine-tuning on math task; (2) Preservation of safety when fine-tuning on benign data; (3) Preservation of safety when fine-tuning on poisoned data.

\begin{table*}[h]
  \centering
  \resizebox{0.99\textwidth}{!}{
  \begin{tabular}{|l|c|c|cccc|ccc|}
    \hline
    \multicolumn{2}{|l|}{\textbf{Method}} & \#\textbf{Params} & \textbf{TriviaQA}$\uparrow$ & \textbf{NQ-open}$\uparrow$ & \textbf{WebQS}$\uparrow$ & \textbf{Avg}$\uparrow$ & \textbf{GSM8k}$\uparrow$ & \textbf{MATH}$\uparrow$ & \textbf{Avg}$\uparrow$ \\
    \hline
    \multicolumn{2}{|l|}{Llama-2-7b} & - & 52.52 & 18.86 & 5.86 & 25.75 & - & - & -  \\
    \hline
    \multicolumn{2}{|l|}{Full fine-tuning} & 6738M & 47.42 & 4.16 & 6.64 & 19.41 & 50.27 & 6.94 & 28.60  \\
    \multicolumn{2}{|l|}{LoRA } & 320M & 46.81 & 1.05 & 7.04 & 18.30 & 41.77 & 5.46 & 23.62   \\
    \multicolumn{2}{|l|}{PiSSA} & 320M & 47.44 & 3.32 & 6.84 & 19.20 & 51.63 & 7.70 & 29.67   \\
    \multicolumn{2}{|l|}{CorDA IPA} & 320M & 30.20 & 9.83 & 5.41 & 15.15 & 51.40 & 8.34 & 29.87  \\
    \multicolumn{2}{|l|}{CorDA KPA} & 320M & 46.21 & \textbf{10.64} & \textbf{7.33} & 21.39 & 45.03 & 6.54 & 25.79   \\
    \hline
    \multirow{3}{*}{\makecell{SC-LoRA}}
    & $\beta=0$ & 320M & 44.26 & 5.18 & 7.19 & 18.88 & \textbf{53.53} & \textbf{8.98} & \textbf{31.25}  \\
    & $\beta=0.5$ & 320M & 48.91 & 7.70 & 6.89 & 21.17 & 53.37 & 8.62 & 31.00  \\
    & $\beta=0.8$ & 320M & \textbf{50.52} & \textbf{10.64} & 7.04 & \textbf{22.73} & 52.46 & 7.62 & 30.04  \\
    \hline
  \end{tabular}}
  \caption{Results of world knowledge preservation and math ability after fine-tuning on MetaMATH. }
  \label{tab:world knowledge, math, result}
\end{table*}

\subsection{World Knowledge Preservation}\label{subsec: world knowledge}

Pre-trained LLMs also have other pre-trained knowledge that is easy to lose after fine-tuning on downstream tasks, such as world knowledge \cite{CorDA}.
In this setting, we aim to preserve the intrinsic world knowledge (e.g., common sense) within the pre-trained LLM while providing efficient fine-tuning on downstream tasks. We fine-tune the Llama-2-7b model \citep{llama2} on math task and evaluate its math ability (utility) and world knowledge performance. We train on 100000 samples of MetaMathQA \citep{yu2024metamath} for 1 epoch and evaluate its math ability on GSM8k \citep{cobbe2021trainingverifierssolvemath} and MATH \citep{yu2024metamath} validation sets. World knowledge is evaluated by the exact matching score on TriviaQA \citep{joshi-etal-2017-triviaqa}, NQ-open \citep{lee-etal-2019-latent}, and WebQS \citep{berant-etal-2013-semantic} through Evaluation-Harness \cite{eval-harness}. We select 256 random samples from NQ-open as world knowledge samples used for the initialization of SC-LoRA and CorDA KPA, and 256 random samples from MetaMathQA as fine-tuning dataset for initializing SC-LoRA and CorDA IPA. Note that samples used in initialization are separate from those in evaluation. 

As shown in Table \ref{tab:world knowledge, math, result}, the results of full fine-tuning and LoRA show the degradation on world knowledge when fine-tuning on downstream task MetaMathQA. SC-LoRA achieves best math ability (surpassing full fine-tuning), and preserves world knowledge relatively well. When $\beta=0.8$, it surpasses all baselines on both utility and knowledge preservation. Also, from the results of SC-LoRA, we can see a clear trend when increasing $\beta$, that the knowledge preservation ability is increasing while the utility is decreasing, which aligns with our design methodology for $\beta$ in Section \ref{sec:method}. More details will be shown in Section \ref{subsec:analyze-beta} to analyze this trend.

\begin{table*}[t]
  \centering
  \begin{tabular}{|l|c|c|cc|c|}
    \hline
    \multicolumn{2}{|l|}{\textbf{Method}} & \#\textbf{Params} & \textbf{HS}$\downarrow$ & \textbf{HR(\%)}$\downarrow$ & \textbf{Utility}$\uparrow$ \\
    \hline
    \multicolumn{2}{|l|}{Llama-2-7b-Chat} & - & 1.100 & 1.212 & 24.13  \\
    \hline
    \multicolumn{2}{|l|}{Full fine-tuning} & 6738M & 1.364 & 5.455 & 51.41 \\
    \multicolumn{2}{|l|}{LoRA} & 320M & 1.176 & 2.424 & 50.32  \\
    \multicolumn{2}{|l|}{PiSSA} & 320M & 1.252 & 4.242 & 51.87  \\
    \multicolumn{2}{|l|}{CorDA IPA} & 320M & 1.209 & 3.333 & 44.61  \\
    \multicolumn{2}{|l|}{CorDA KPA} & 320M & 1.106 & 0.606 & 50.89  \\
    \hline
    \multirow{3}{*}{SC-LoRA}
    & $\beta=0.5$ & 320M & 1.161 & 1.818 & \textbf{52.54}  \\
    & $\beta=0.7$ & 320M & 1.148 & 1.818 & 52.07  \\
    & $\beta=0.9$ & 320M & \textbf{1.097} & \textbf{0.000} & 51.67  \\
    \hline
  \end{tabular}  
  \caption{Results of Safety preservation and fine-tuning performance when training on benign dataset Samsum. \#Params is the number of trainable parameters. HS and HR denote harmfulness score and harmfulness rate respectively.}
  \label{tab:safety, samsum, result} 
\end{table*}

\subsection{Safety Preservation on Benign Finetuning}\label{subsec: benign finetuning}

\citet{benign_finetune_attack_hex_phi} has shown that fine-tuning on benign data can compromise the safety of aligned LLMs. In this setting, we aim to preserve the safety of aligned LLM while providing efficient fine-tuning on downstream tasks. Following the experimental settings by \citet{shallow_deep}, we fine-tune Llama-2-7b-Chat model with safety alignment \citep{llama2} on Samsum \citep{gliwa-etal-2019-samsum} for 1 epoch. Samsum is a dataset for conversation summarization task, containing 14732 training samples and 819 testing samples. 

To initialize our SC-LoRA model, we randomly select 256 samples from training set of Samsum ($\mathcal{D}_+$) to compute covariance matrix $\mathrm{Cov}_+$ for each linear layer, then use 256 harmful-question\&refusal-answer pairs (as the safety dataset $\mathcal{D}_-$) provided by \citet{shallow_deep} to compute $\mathrm{Cov}_-$. 
These two collections of samples are also used to compute the covariance matrices of CorDA IPA and CorDA KPA respectively. 
For better comparability, we tune the learning rate of all LoRA-based baselines to 2e-5, 5e-5 and 1e-4, and pick the best result among all learning rates. The learning rate for SC-LoRA is fixed to 2e-5. Full results of learning rate tuning can be seen in Appendix \ref{sec:pareto} Figure \ref{fig:samsum_pareto}.

For utility evaluation, we employ the standard ROUGE-1 score \citep{lin-2004-rouge} for testing set of Samsum. 
For safety evaluation, we let the fine-tuned models to generate answers for 330 malicious questions provided by \citet{benign_finetune_attack_hex_phi} (distinct from malicious questions for initialization) and employ DeepSeek-V3 \citep{deepseekv3} API to judge the harmfulness, assigning each answer an integer score from 1 (safe) to 5 (most harmful). We report the average score as \textbf{harmfulness score} of the model and the fraction of maximum-risk responses (score = 5) as \textbf{harmfulness rate}. Lower values for both metrics indicate stronger safety of the model. 

As shown in Table \ref{tab:safety, samsum, result}, SC-LoRA achieves high utility, even surpassing full fine-tuning on Samsum dataset when $\beta=0.5$. At the same time, SC-LoRA shows almost no safety degradation compared to the model before fine-tuning, while all baselines except CorDA KPA present notable safety degradation, since they are not designed for knowledge preservation. However, the utilities of all fine-tuning methods (except for CorDA IPA) are generally close. 
We hypothesize that the task of summarization is quite simple, so training for only 1 epoch is enough for utility convergence. 
Also, the results of SC-LoRA shows that when $\beta$ is increasing, the safety preservation becomes better while utility is decreasing. This aligns with our design of $\beta$ to balance the trade-off. 

\begin{table*}[t]
  \centering
  \begin{tabular}{|l|c|c|cc|c|}
    \hline
    \multicolumn{2}{|l|}{\textbf{Method}} & \#\textbf{Params} & \textbf{HS}$\downarrow$ & \textbf{HR(\%)}$\downarrow$ & \textbf{Utility}$\uparrow$ \\
    \hline
    \multicolumn{2}{|l|}{Llama-2-7b-Chat} & - & 1.100 & 1.212 & -  \\
    \hline
    \multicolumn{2}{|l|}{Full fine-tuning} & 6738M & 2.248 & 23.94 & 41.47 \\
    \multicolumn{2}{|l|}{LoRA} & 320M & 2.276 & 23.64 & 37.68  \\
    \multicolumn{2}{|l|}{PiSSA} & 320M & 2.379 & 29.39 & 41.77  \\
    \multicolumn{2}{|l|}{CorDA IPA} & 320M & 4.239 & 67.27 & 43.75  \\
    \multicolumn{2}{|l|}{CorDA KPA} & 320M & 1.127 & \textbf{1.212} & 40.33  \\
    \hline
    \multirow{3}{*}{SC-LoRA} 
    & $\beta=0.5$ & 320M & 1.630 & 10.91 & \textbf{45.56}  \\
    & $\beta=0.7$ & 320M & 1.224 & 3.030 & 45.26  \\
    & $\beta=0.9$ & 320M & 1.136 & \textbf{1.212} & 45.26  \\
    \hline
  \end{tabular} 
  \caption{Results of safety preservation and fine-tuning performance when training on poisoned dataset MetaMathQA with 1\% malicious question-answer pairs. }
  \label{tab:safety, math poisoned, result}
\end{table*}

\begin{table*}[t]
\centering
\begin{tabular}{|l|c|c|c|c|}
\hline
\textbf{Finetuning Task} & \textbf{\#sample} & \textbf{HS}$\downarrow$ & \textbf{HR(\%)} $\downarrow$ & \textbf{Utility}$\uparrow$ \\
\hline
\multirow{4}{*}{\textbf{Data poisoning}} & 256 & 1.136 & 1.212 & 45.26 \\
 & 128    & 1.155 & 2.121 & 45.64 \\
 & 64     & 1.200 & 2.727 & 46.10 \\
 & 32     & 1.864 & 1.616 & 45.72 \\
\hline
\multirow{4}{*}{\textbf{Benign finetuning}} & 256 & 1.097 & 0.000 & 51.67 \\
 & 128    & 1.118 & 1.515 & 51.91 \\
 & 64     & 1.145 & 1.515 & 52.46 \\
 & 32     & 1.158 & 1.515 & 52.42 \\
\hline
\end{tabular}
\caption{Sensitivity to the Number of Initialization Samples.}
\label{tab:num_init_samples}
\end{table*}

\subsection{Safety Preservation on Data Poisoning Attack}\label{subsec: Safety Preservation on Data Poisoning Attack}

Harmful data injection is a common attack method to degrade the safety of LLMs during fine-tuning \citep{huang2024booster,huang2024lazy, huang2024vaccine}. In this experiment, we aim to preserve safety in poisoned data scenarios. To construct the poisoned dataset, we first take 25600 data samples from training set of MetaMathQA \citep{yu2024metamath}, then replace 1\% of the data by harmful question-answer pairs provided by \citep{benign_finetune_attack_hex_phi}. We train each method for 1 epoch on the poisoned dataset. 
For the initialization of SC-LoRA and CorDA IPA, we use 256 samples from training set of MetaMathQA. The safety samples used for the initialization of SC-LoRA and CorDA KPA are the same with the previous experiment (Section \ref{subsec: benign finetuning}). 

For utility evaluation, we compute the answer accuracy on the validation set of GSM8k \citep{cobbe2021trainingverifierssolvemath}. Safety evaluation follows the setting in the previous section \ref{subsec: benign finetuning}. 
For better comparability, we tune the learning rate of all LoRA-based baselines to 2e-5, 5e-5 and 1e-4, and pick the best result among all learning rates. The learning rate for SC-LoRA is fixed to 2e-5. Full results of learning rate tuning can be seen in Appendix \ref{sec:pareto} Figure \ref{fig:math_mix_harmful_pareto}.

From the results in Table \ref{tab:safety, math poisoned, result}, we can observe that the data points exhibit a wider spread among these methods, both in utility and safety metric. Compared to the original model, 
SC-LoRA ($\beta=0.9$) exhibits almost no safety degradation, and 
achieves best utility, even surpassing full fine-tuning by 3.79 points. When increasing the learning rate, LoRA shows a sharp decline in safety alignment while math ability is increasing. LoRA (lr=2e-5) and CorDA KPA, though preserving safety well, are insufficient in fine-tuning performance compared to our method. PiSSA and CorDA IPA, though showing their capacity in better fine-tuning, heavily degrades the safety of the model. This again shows the potential of our method to enhance the utility of the model and preserve safety at the same time, even when the fine-tuning dataset contains a small fraction of harmful content.  Also, the utility and safety of SC-LoRA follows the same trend as in fine-tuning on benign data when $\beta$ is increasing, supporting the sedign of our method.

\begin{figure*}[h]
\centering
\begin{subfigure}{0.49\linewidth}
    \includegraphics[width=\linewidth]{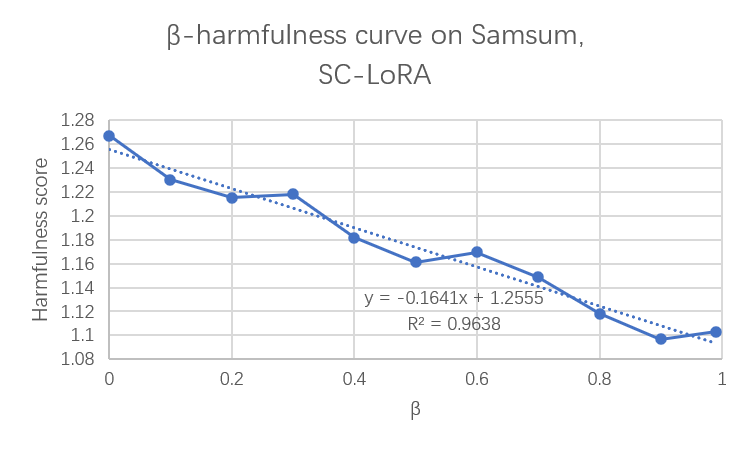} 
\end{subfigure}
\begin{subfigure}{0.49\linewidth}
    \includegraphics[width=\linewidth]{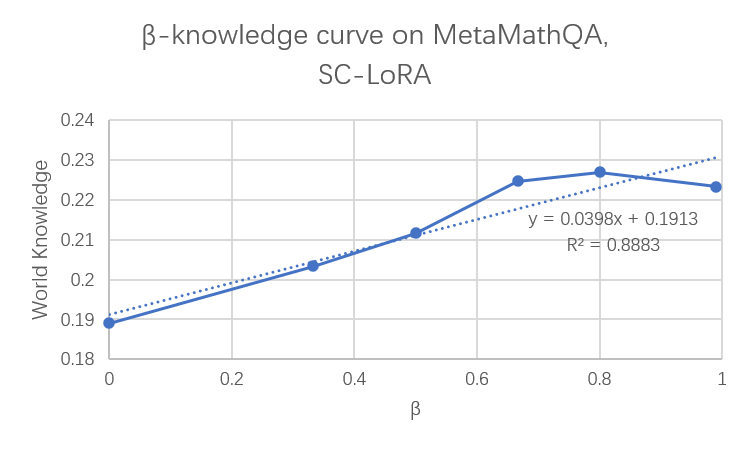} 
\end{subfigure}
\caption{Relations between $\beta$ and knowledge preservation performance. The experiment setting of the left figure is described in Section \ref{subsec: benign finetuning}, while that of the right figure is described in Section \ref{subsec: world knowledge}. Lower harmfulness score or higher world knowledge score indicates better performance on knowledge preservation. }
\label{fig: trend, knowledge}
\end{figure*}

\begin{figure*}[h]
\centering
\begin{subfigure}{0.49\linewidth}
    \includegraphics[width=\linewidth]{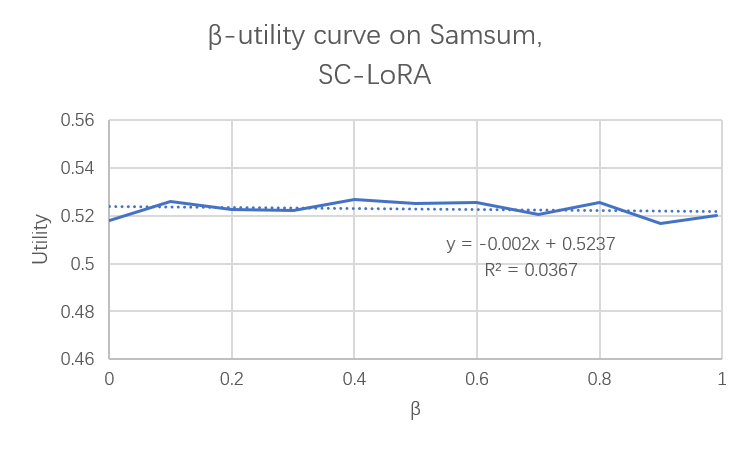} 
\end{subfigure}
\begin{subfigure}{0.49\linewidth}
    \includegraphics[width=\linewidth]{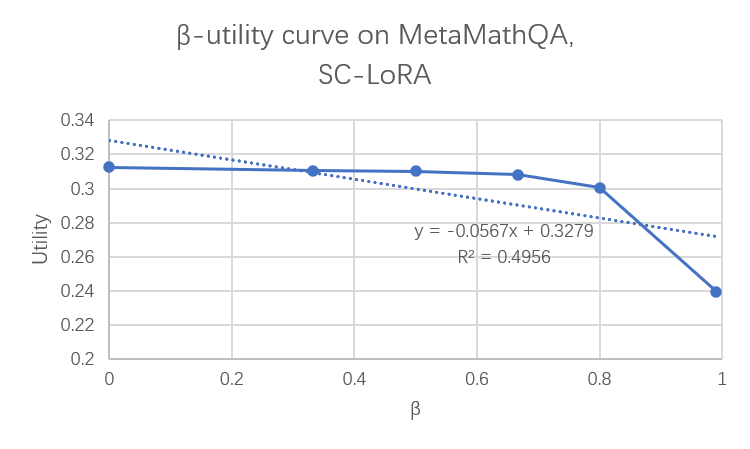} 
\end{subfigure}
\caption{Relations between $\beta$ and fine-tuning  performance. The experiment setting of the left figure is described in Section \ref{subsec: benign finetuning}, while that of the right figure is conducted in Section \ref{subsec: world knowledge}. The right figure shows clear monotonicity with $\beta$, while such trend does not occur in the left figure. }

\label{fig: trend, utility}
\end{figure*}

\subsection{Ablation: Experimental Analysis on the Functionality of Hyper-Parameter $\beta$}
\label{subsec:analyze-beta}

As explained in Section \ref{sec:method}, the value of $\beta$ balance the trade-off between knowledge preservation and fine-tuning efficiency. Intuitively, when increasing $\beta$, there exists a trend that the fine-tuning performance will drop and the knowledge preservation ability will increase.
While we have observed this trend in the previous section, we illustrate the trend more explicitly in Figure \ref{fig: trend, knowledge} and \ref{fig: trend, utility}. In Figure \ref{fig: trend, knowledge}, both two curves shows knowledge preservation improvement when $\beta$ is increasing: one for safety increasing, and the other for world knowledge preservation improvement. In Figure \ref{fig: trend, utility}, the math ability decreases when $\beta$ is increasing, aligning with our expectations. The utility on Samsum, however, does not show evident trend as $\beta$ varies, but fluctuating around 0.52. We hypothesize that the task of summerization is quite simple, so whatever the value of $\beta$, it is sufficient for utility convergence during fine-tuning.

These trends give experimental support to our method design, that by adjusting $\beta$ we can balance the trade-off. Interestingly, a linear relationship was observed between $\beta$ values and knowledge preservation in some experimental settings.

\subsection{Ablation: Number of Samples for Initialization}
\label{sec:num_init_sample}

A key aspect of our data-driven initialization is the number of samples used to estimate the covariance matrices. A robust method should not require an impractically large number of samples to be effective. In this section, we analyze the sensitivity of SC-LoRA's performance to the number of initialization samples.

We conduct this analysis under two scenarios described earlier: data poisoning (Section \ref{subsec: Safety Preservation on Data Poisoning Attack}) and benign fine-tuning (Section \ref{subsec: benign finetuning}). For both scenarios, we vary the number of samples for both the fine-tuning task ($D_+$) and the preserved safety task ($D_-$) from $32$ to $256$, keeping all other hyperparameters constant ($\beta=0.9$, learning rate $ 2e-5 $). The results are presented in Table \ref{tab:num_init_samples}. From the table, we can draw several key conclusions:
\textbf{(1) Safety Preservation Improves with More Samples}. 
This is expected, as a larger sample set provides a more accurate and stable estimate of the true covariance of the data distributions, allowing our method to better identify and constrain the safety-critical subspace.
\textbf{(2) Utility Remains Stable and Insensitive to Sample Size}. 
For both scenarios, the utility score only fluctuates within a range of $1\%$. A possible reason is that even a small number of samples (like $32$) is sufficient to capture the principal directions of the fine-tuning task ($T_+$). Hence adding more samples may not significantly alter the subspace chosen for adaptation, leading to consistent performance on the downstream task.

While more samples are generally better for safety, the results suggest that for practical applications, using $128$ to $256$ samples provides a reliable balance between computational cost during initialization and robust performance. This is a practical and accessible number, reinforcing the efficiency of our approach.

\section{Conclusion}
\label{sec:conclusion}

Aimed to balance the trade-off between efficient fine-tuning and knowledge preservation, this paper presents a data-driven LoRA initialization that utilizes the subspace constrain, in order to strengthen the target knowledge while downgrading its influence on preserved knowledge. Theoretical analysis are provided to support our method, including the choice of subspace and the initialization setting. 
We conduct extensive experiments regrading safety preservation and world knowledge preservation, during fine-tuning on various downstream tasks such as math and summarization. The results of experiments strongly demonstrate that our method can not only promote fine-tuning performance on downstream tasks, but also preserve the intrinsic knowledge stored in pre-trained model, surpassing contemporary LoRA initialization methods.

\section{Limitations}
\label{sec:limitations}

First, SC-LoRA is just a LoRA initialization method, and does not strongly constrain the updates during fine-tuning process. Hence after fine-tuning on more complex tasks and with more steps, the knowledge preservation ability can also drop (see the preservation drop of NQ-open in Table \ref{tab:world knowledge, math, result} for example). 
Second, its application on preserving other types of knowledge remains unexplored. Future work may consider applying SC-LoRA to preserving multimodal large language model's performance on pre-training tasks \citep{pmlr-v234-zhai24a} or large language model's reasoning ability. 

These aspects provide promising directions for future researches.

\section*{Acknowledgments}

We are grateful to Prof. Jingzhao Zhang from Tsinghua University for helpful discussions and constructive comments on this work.

\bibliography{custom}

\clearpage

\appendix


\section{Proof Details}
\label{sec:def-proof}

\begin{figure*}[h]
    \centering
    \includegraphics[width=0.9\textwidth]{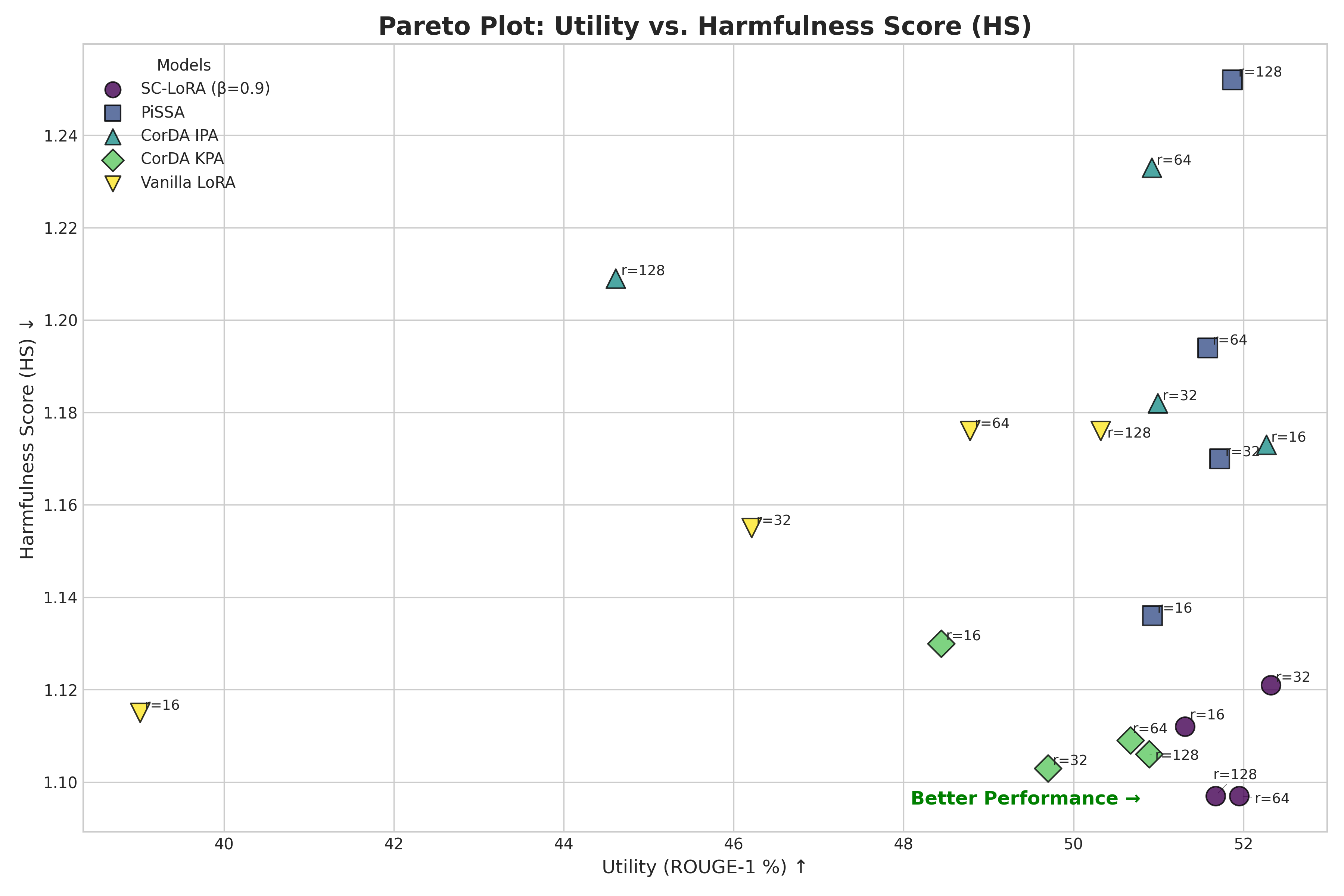}
    \caption{Sensitivity to the rank $r$ on Samsum.}
    \label{fig:rank}
\end{figure*}

\subsection{Proof for Theorem \ref{thm:reward}}\label{proof:reward}

\begin{proof}

Suppose for some subspace $S \subset \mathbb{R}^{d}$ (ignore the subscript of $d_{\mathrm{out}}$ for simplicity) of dimension $r$, there exists an orthonormal basis $\left\{ v_i \right\}_{i\in[r]}$ that spans $S$, that is $S = \mathrm{span}\left(\left\{ v_i \right\}_{i\in[r]}\right)$. 

For simplicity, denote 

\begin{equation}
  \tilde{I}_r = \sum_{i\in[r]} v_i v_i^\top,
\end{equation}

then the following equality holds: 

\begin{equation}
  \begin{aligned}
    \tilde{I}_r^\top \tilde{I}_r &= \sum_{i\in[r]} \sum_{j\in[r]} v_i v_i^\top v_j v_j^\top \\ &= \sum_{i\in[r]} \sum_{j\in[r]} v_i \langle v_i, v_j\rangle v_j^\top \\
    &= \sum_{i\in[r]} \sum_{j\in[r]} \delta_{ij}v_i v_j^\top \\ 
    &= \sum_{i\in[r]} v_i v_i^\top = \tilde{I}_r. 
  \end{aligned}
\end{equation}

From property of projection, 

\begin{equation}
  \begin{aligned}
    \Pi_S\left(X_\pm\right) &= \sum_{i=1}^{r} \langle X_\pm, v_i \rangle v_i = \sum_{i=1}^{r} v_i v_i^\top X_\pm \\ &= \left(\sum_{i=1}^{r} v_i v_i^\top \right) X_\pm = \tilde{I}_r X_\pm.
  \end{aligned}
\end{equation}

Thus 

\begin{equation}
  \begin{aligned}
    &\mathbb{E}_{X_\pm\sim \mathcal{P_\pm}}\left[\left\|\Pi_S\left(X_\pm\right)\right\|_2^2\right]
    \\
    =& \mathbb{E}_{X_\pm\sim \mathcal{P_\pm}}\left[\left\|\tilde{I}_r X_\pm\right\|_2^2\right] \\
    =& \mathbb{E}_{X_{\pm} \sim \mathcal{P}_{\pm}}\left[ \mathrm{tr}\left(X_\pm^\top \tilde{I}_r^\top \tilde{I}_r X_\pm \right)\right] \\
    =& \mathbb{E}_{X_{\pm} \sim \mathcal{P}_{\pm}}\left[ \mathrm{tr}\left(X_\pm^\top \tilde{I}_r X_\pm \right)\right] \\
    =& \mathbb{E}_{X_{\pm} \sim \mathcal{P}_{\pm}}\left[ \mathrm{tr}\left(\tilde{I}_r X_\pm X_\pm^\top\right)\right] \\
    =& \mathrm{tr}\left(\tilde{I}_r \mathbb{E}_{X_{\pm} \sim \mathcal{P}_{\pm}} \left[X_\pm X_\pm^\top\right]\right) \\ =& \mathrm{tr}\left(\tilde{I}_r \mathrm{Cov}_\pm \right) . 
  \end{aligned}
\end{equation}

Suppose the spectral decomposition of $(1-\beta)\mathrm{Cov}(X_+) - \beta\mathrm{Cov}(X_-)$ is $Q \Sigma Q^\top$, where $Q = (q_1\, q_2\, \cdots \, q_d)$, $\Sigma$ is diagonal with eigenvalues sorted in descending order. Then we have

\begin{equation}
  \begin{aligned}
  R(S) &= (1-\beta)\mathbb{E}_{X_{+} \sim \mathcal{P}_{+}} \left[ \left\| \Pi_S\left(X_+\right) \right\|_2^2 \right] \\ &- \beta \mathbb{E}_{X_{-} \sim \mathcal{P}_{-}} \left[ \left\|\Pi_S(X_-)\right\|_2^2 \right] \\
  &= (1-\beta)\mathrm{tr}\left(\tilde{I}_r \mathrm{Cov}_{+} \right) - \beta  \mathrm{tr}\left( \tilde{I}_r \mathrm{Cov}_{-} \right) \\ 
  &= \mathrm{tr}\left( \tilde{I}_r \Delta\mathrm{Cov} \right) \\
  &= \sum_{i\in[r]}\mathrm{tr}\left( v_i v_i^\top Q \Sigma Q^\top \right) \\
  &= \sum_{i\in[r]} v_i^\top Q \Sigma Q^\top v_i .
  \end{aligned}
\end{equation}

Extend $\left\{ v_i \right\}_{i\in[r]}$ to a complete orthonormal basis $\left\{ v_i \right\}_{i\in[d]}$ for $\mathbb{R}^d$, and denote $u_i = Q_i^\top v_i$. Since $Q$ is an orthogonal matrix, $\{u_i\}_{i\in [d]}$ is also an orthonormal basis for $\mathbb{R}^d$. From Ky Fan's theorem on eigenvalues, $\max \left( \sum_{i\in[r]} v_i^\top Q \Sigma Q^\top v_i\right) = \sum_{i\in [r]} \Sigma_{ii}$, and one can easily verify that the condition above achieves the maximum. 

For the if and only if part (adding the condition of eigenvalue gap): suppose $U = (u_1\, u_2\, \cdots \, u_d)^\top$ as an orthogonal matrix, then 

\begin{equation}
  \begin{aligned}
  R(\{v_i\}_{i\in[r]}) &= \sum_{i\in[r]} u_i^\top \Sigma u_i \\ &= \sum_{i\in[r]}\sum_{j=1}^{d} \Sigma_{jj} U_{ij}^2 \\ &= \sum_{j=1}^{d} \left(\Sigma_{jj} \sum_{i\in[r]} U_{ij}^2 \right).
  \end{aligned}
\end{equation}

From the property of orthogonal matrix, $\sum_{i\in[r]} U_{ij}^2 \le1$ and $\sum_{j=1}^{d} \sum_{i\in[r]} U_{ij}^2 = r$, then to maximize $R$, from the additional assumption we need $\sum_{i\in[r]} U_{ij}^2 = \begin{cases}1, &1 \le j\le r \\ 0, &r+1 \le j \le d \end{cases}$, which is equivalent to 

\begin{equation}
  U_{1:r,1:d}^\top U_{1:r,1:d} = \begin{pmatrix}I_r & O \\ O & O \end{pmatrix}.
\end{equation}

From $U_{1:r,1:d} = (v_1\,v_2\,\cdots \, v_r)^\top Q$, we know that this is also equivalent to 

\begin{equation}
  (v_1\,v_2\,\cdots \, v_r) (v_1\,v_2\,\cdots \, v_r)^\top = Q \begin{pmatrix}I_r & O \\ O & O \end{pmatrix} Q^\top ,
\end{equation}

which is also written as

\begin{equation}
  \sum_{i=1}^{r} v_i v_i^\top = \sum_{i=1}^{r} q_i q_i^\top .
\end{equation}

Indicating $S = \mathrm{span}\left(\left\{q_i\right\}_{i\in[r]}\right)$. 

\end{proof}

\subsection{Proof of Theorem \ref{thm:init-of-BA}}
\begin{proof}
\label{proof:init-of-BA}
Denote $Q_r = (q_1\,q_2\,\cdots\,q_r)$. 

Since $\{q_i\}_{i\in[r]}$ is a orthonormal basis that spans $S$, from definition of orthogonal projection we have 

\begin{equation}
  \Pi_S(h) = \sum_{i=1}^{r} q_i q_i^\top h = Q_r Q_r^\top h. 
\end{equation}

Thus $\forall x \in \mathbb{R}^{d_{\mathrm{in}}}$, we have
\begin{equation}
  B_{\mathrm{init}} A_{\mathrm{init}} x = Q_r Q_r^\top W_0 x = Q_r Q_r^\top h = \Pi_S(h),
\end{equation}

which completes the proof. 

\end{proof}

\section{Ablation Study on Low Rank $r$}

This section studies the sensitivity to the rank $r$ under experiment \ref{subsec: benign finetuning} (Benign Finetuning on Samsum). The results are plotted in Figure \ref{fig:rank}. Generally, SC-LoRA with $\beta=0.9$ achieves almost the best utility with small safety degradation on ranks from 16 to 128.

\section{Learning Rate Fine-tuning}
\label{sec:pareto}

\begin{figure*}[h]
    \centering
    \includegraphics[width=0.9\linewidth]{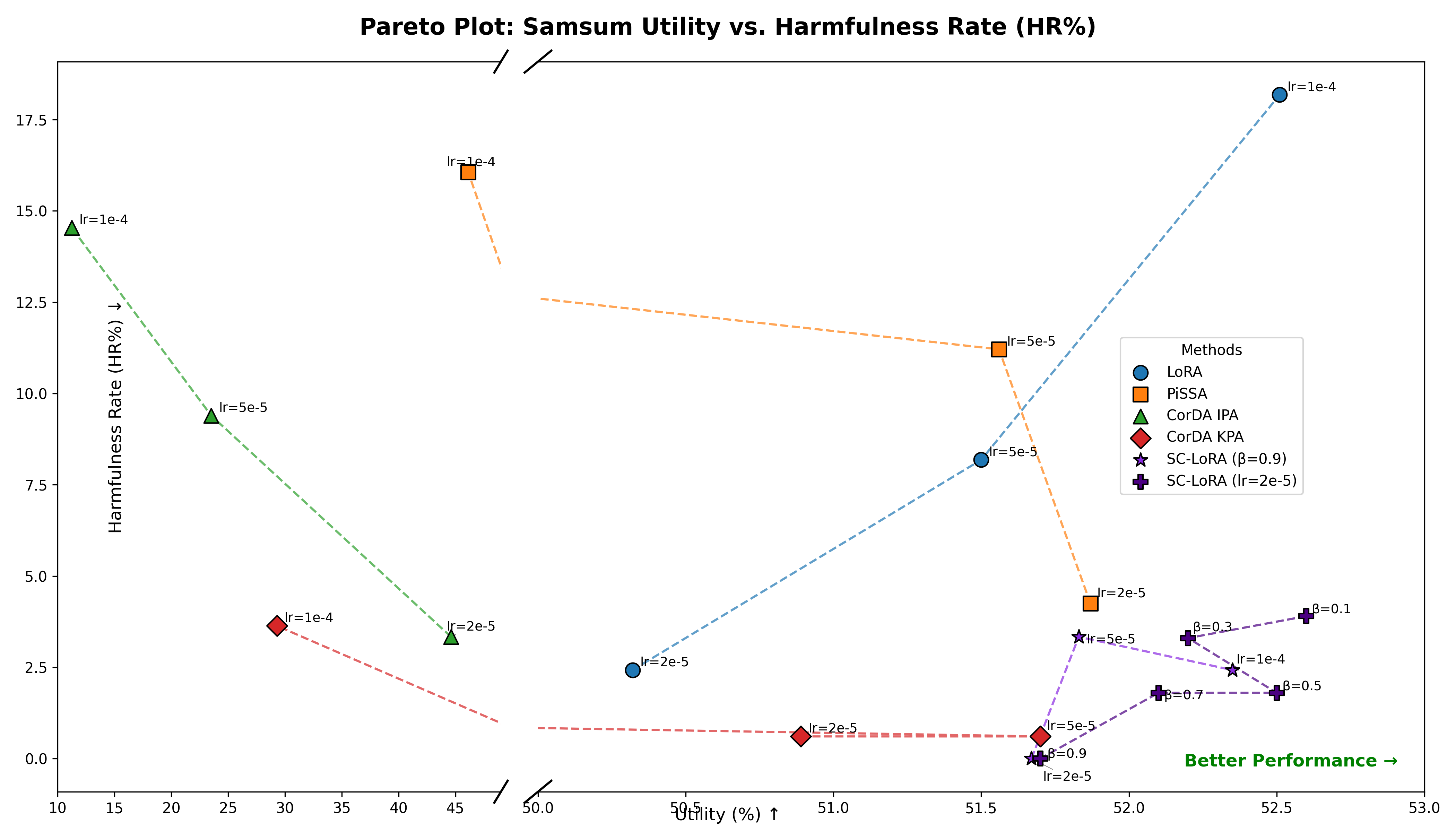}
    \caption{Pareto plot of learning rate fine-tuning on task Samsum.}
    \label{fig:samsum_pareto}
\end{figure*}

\begin{figure*}[h]
    \centering
    \includegraphics[width=0.9\linewidth]{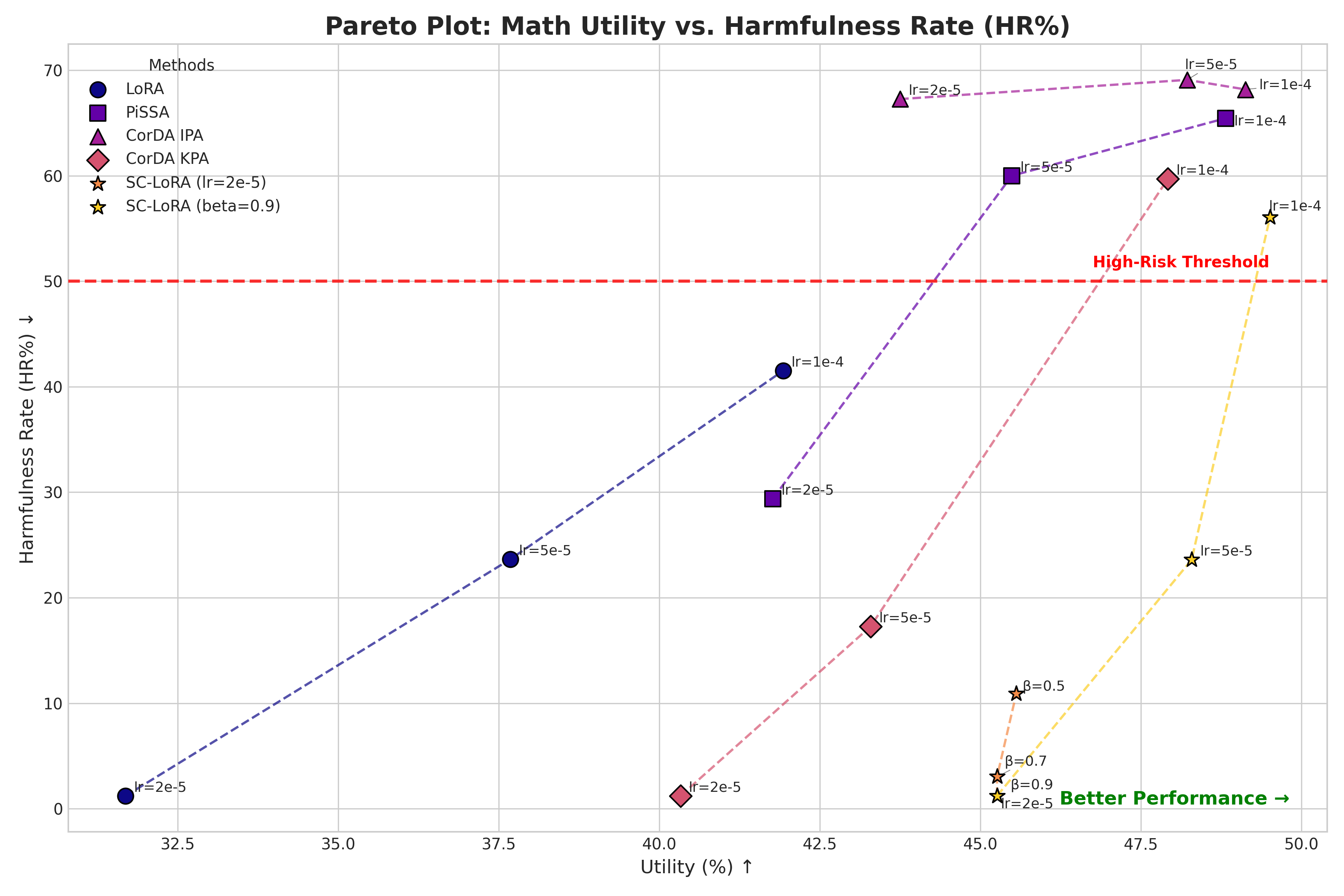}
    \caption{Pareto plot of learning rate fine-tuning on task poisoned math.}
    \label{fig:math_mix_harmful_pareto}
\end{figure*}

This section presents the Pareto plots of learning rate fine-tuning and $\beta$ fine-tuning. See Figure \ref{fig:samsum_pareto} and Figure \ref{fig:math_mix_harmful_pareto}. The curve is closer to the bottom right, the safer response and better performance the model presents. From the Pareto plot, our SC-LoRA curve can best balance safety preservation and utility learning compared to other methods.

\begin{table*}[h]
\centering
\begin{tabular}{|l|c|c|c|c|}
\hline
\textbf{Finetuning Task} & \textbf{\#samples} & \textbf{HS}$\downarrow$ & \textbf{HR(\%)} $\downarrow$ & \textbf{Utility}$\uparrow$ \\
\hline
\multirow{4}{*}{\textbf{Data poisoning}} & 256 & 1.136 & 1.212 & 45.26 \\
 & 128    & 1.155 & 2.121 & 45.64 \\
 & 64     & 1.200 & 2.727 & 46.10 \\
 & 32     & 1.864 & 1.616 & 45.72 \\
\hline
\multirow{4}{*}{\textbf{Benign finetuning}} & 256 & 1.097 & 0.000 & 51.67 \\
 & 128    & 1.118 & 1.515 & 51.91 \\
 & 64     & 1.145 & 1.515 & 52.46 \\
 & 32     & 1.158 & 1.515 & 52.42 \\
\hline
\end{tabular}
\caption{Sensitivity to the Number of Initialization Samples.}
\label{tab:num_init_samples}
\end{table*}

\section{Number of Samples in Initialization}

This section compares the finetuning performance on different number of samples used to estimate the covariance matrices for SC-LoRA's initialization. We compare under the experiment \ref{subsec: Safety Preservation on Data Poisoning Attack} (Safety Preservation on Data Poisoning Attack) and \ref{subsec: benign finetuning} (Safety Preservation on Benign Finetuning). 

The results shows that performance remains relatively stable and effective for both tasks using 128 or more samples. Number of samples within 64 causes degradation in preservation. 

\section{Ablation: Number of Samples for Initialization}
\label{sec:num_init_sample}

A key aspect of our data-driven initialization is the number of samples used to estimate the covariance matrices. A robust method should not require an impractically large number of samples to be effective. In this section, we analyze the sensitivity of SC-LoRA's performance to the number of initialization samples.

We conduct this analysis under two scenarios described earlier: data poisoning (Section \ref{subsec: Safety Preservation on Data Poisoning Attack}) and benign fine-tuning (Section \ref{subsec: benign finetuning}). For both scenarios, we vary the number of samples for both the fine-tuning task ($D_+$) and the preserved safety task ($D_-$) from $32$ to $256$, keeping all other hyperparameters constant ($\beta=0.9$). The results are presented in Table \ref{tab:num_init_samples}. From the table, we can draw several key conclusions:
\textbf{(1) Safety Preservation Improves with More Samples}. 
This is expected, as a larger sample set provides a more accurate and stable estimate of the true covariance of the data distributions, allowing our method to better identify and constrain the safety-critical subspace.
\textbf{(2) Utility Remains Stable and Insensitive to Sample Size}. 
For both scenarios, the utility score only fluctuates within a range of $1\%$. A possible reason is that even a small number of samples (like $32$) is sufficient to capture the principal directions of the fine-tuning task ($T_+$). Hence adding more samples may not significantly alter the subspace chosen for adaptation, leading to consistent performance on the downstream task.

While more samples are generally better for safety, the results suggest that for practical applications, using $128$ to $256$ samples provides a reliable balance between computational cost during initialization and robust performance. This is a practical and accessible number, reinforcing the efficiency of our approach.

\section{Selection of hyperparameter $\beta$ in practice}\label{numerical instability in sparse sample setting}

When the sample size of initialization is much larger than the output activation dimension, i.e. $\min(|\mathcal{D}_+|L,|\mathcal{D}_-|L) \gg d_{\mathrm{out}}$, setting $\beta \in [0,1]$ causes no issue. However when samples are sparse (specifically, when the number of negative-task sample $B_- < (d_{\mathrm{out}}-r)/L$, setting $\beta=1$ introduces multiple valid solutions in the spectral decomposition step due to high-dimensional freedom in the null space of $\mathrm{Cov}_-$. Mathematically, the rank of $\mathrm{Cov}_-$ is at most $B_- L$, resulting in a null space of dimension $d_{\mathrm{out}} - \mathrm{rank}(\mathrm{Cov}_-) \ge d_{\mathrm{out}} - B_-L > r$. Consequently, \textbf{any arbitrary set of $r$ orthonormal vectors} in this null space can satisfy the decomposition criterion, leading to non-unique initialization of parameters $A$ and $B$. 

In practice, $B_- \sim (d_{\mathrm{out}}-r)/L$, the decomposition results may also be affected by data selection and clipping, demonstrated by the drop of utility for $\beta=0.99$ (exhibited in Section \ref{subsec:analyze-beta}). 

To mitigate this instability, we recommend setting $1-\beta$ to a small positive value (rather than exactly zero), for example $\beta = 0.8$ or $0.9$ as discussed in Section \ref{sec:experiments}. This retains the regularization from $\mathrm{Cov}_+$ in the objective function, which constrains the null space ambiguity and stabilizes the spectral decomposition, empirically improves fine-tuning performance.

\end{document}